# Predicting Movie Production Years through Facial Recognition of Actors with Machine Learning

**Asraa Muayed Abdalah[1*]**   **Noor Redha Alkazaz[1]**

[1]Department of Computer Science, College of Science for Women, University of Baghdad, Baghdad, Iraq.
*Corresponding author: asraa.m@csw.uobaghdad.edu.iq
E-mail Address: noor.alkazaz@csw.uobaghdad.edu.iq

**ABSTRACT:** This study used different machine learning algorithms to identify actors and extract the age of actors from images taken randomly from movies. The use of images taken from Arab movies includes challenges such as non-uniform lighting, different and multiple poses for the actors and multiple elements with the actor or a group of actors. Additionally, the use of make-up, wigs, beards, and wearing different accessories and costumes made it difficult for the system to identify the personality of the same actor. Different age stages were chosen for the same actor, such as Nour al-Sharif being twenty years old and forty years old. The Arab Actors Dataset-AAD is a dataset of 574 images taken from various movies, including black and white, and color ones. Also some images represent full scene while others were part of a scene. A comparison among several models for feature extractions. Also a comparison among different machine learning algorithms was made in the classification and prediction stages to find out which algorithm is the best in dealing with such type of images. The study demonstrated the effectiveness of the Logistic Regression model exhibited the best performance compared to other models in the training phase, as evidenced by its AUC, precision, CA and F1score values of 99%, 86%, 85.5% and 84.2% respectively. The results of this research could be used to enhance the precision and dependability of facial recognition systems for diverse applications like movies search engines, movie recommendation systems, and movie genre analysis.

**Keywords:** Artificial Intelligence, Machine Learning Algorithms, Face Recognition, Age Prediction, Naive Bayes (NB), Decision Tree (DT), Support Vector Machine (SVM), and Artificial Neural Network (ANN).

## 1. Introduction:

Face recognition has become an essential research area in computer vision due to its numerous applications in security, surveillance, entertainment, and marketing. Face recognition is the process of identifying an individual based on their facial features, this task is particularly challenging due to the inherent variability in facial appearance caused by factors such as pose, illumination, facial expression, and age-related changes as well as the use of makeup and wigs. The combination of machine learning technique to recognize the same actor in different shapes (because of makeup, wigs depending on the character) can provide valuable insights into understanding human behavior and demographics.

In the film industry, selecting suitable actor for specific role is crucial to the success of a movie. Face features are an important factor to consider when casting an actor, as it can affect a character's authenticity and credibility. However, recognizing an actor from images taken from movies can be difficult due to the use of make-up to alter facial features to suit a specific role. The aim of our paper is to investigate the application of machine learning algorithms in identifying actors and extracting their ages using different Arabic movie images.

Recently, several studies have been conducted to improve the accuracy of face recognition using machine learning algorithms. These algorithms are capable of automatically learning from data and making predictions based on the learned patterns [1]. Naive Bayes (NB), Support Vector Machine (SVM), Decision Tree (DT), and Artificial Neural Network (ANN) are some of the popular machine learning algorithms[2] used for face recognition and age prediction. Prior research, however, frequently concentrated on face-centric datasets with constant illumination, identical resolutions, and little variability in postures. In contrast, our research presents a novel approach that tackles the challenges of identifying actors and extracting their ages from images taken from Arabic movies, where scenes can contain actors in different poses, non-uniform lighting, and encompass various elements beyond just the face.

The central contribution of our work lies in the creation of the Arab Actors Dataset (AAD), a unique collection of 574 images sourced from a variety of Arabic movies. Unlike previous datasets, the AAD captures scenes that feature actors in full, encompassing their facial features, body poses, and the context of the movie. This extensive dataset enables us to investigate the subtleties and difficulties of actor identification and age estimate in a tough and realistic environment.

Our research addresses the limitations of previous studies by employing state-of-the-art machine learning algorithms and comparing them within the context of Arabic movies. We face a number of challenges, including different lighting, various stances, and the inclusion of props like costumes, wigs, and beards that can change an actor's image. Additionally, we take into account age differences between actors. For instance, Nour al-Sharif has portrayed roles between the ages of twenty and forty.

The significance of our research extends beyond the realm of face recognition. The findings of our work have significance for different applications, including movie search engines, movie recommendation systems, and movie genre analysis, by improving the accuracy and dependability of facial recognition systems. Our research clarifies the applicability and efficacy of face recognition systems in a culturally varied and complex setting by tackling the particular difficulties presented by Arabic movies and making use of the extensive AAD dataset.

## 2. **Related Works:**

The secure functioning of an organization relies heavily on the ability to recognize human faces. The field of machine learning has seen a surge of research focused on pattern recognition, with a specific emphasis on human face recognition. In this section exploration of recent developments in this area and examine various algorithm used for this task, along with their respective performances.

Rahul [3] et al. investigated through their review the fundamentals of face recognition through the use of Support Vector Machine (SVM) learning algorithms.

Xinyi [4] et al. conducted a survey that covered different aspects of face recognition technology. The survey started with an overview of face recognition in the Introduction Section. The history of face recognition was discussed in Section 2, while Section 3 focused on explaining the deep learning framework's pipeline. Section 4 provided an in-depth explanation of various face recognition algorithms, such as loss functions, embedding techniques, face recognition with massive IDs, cross-domain, pipeline acceleration, closed-set training, mask face recognition and privacy-preserving. In Section 5, a series of experiments were conducted to evaluate the effect of backbone size and data distribution. The frequently used training and test datasets were presented, along with comparison results, in Section 6. Applications of face recognition were discussed in Section 7, while Section 8 introduced competitions and open-source programs.

Tiago [5] et al. conducted a study in which they expanded upon their prior research conducted in 2014 and published in 2016. The original study explored how different face characteristics affected facial recognition systems. In their latest research, the authors evaluated the effectiveness of modern facial recognition systems against earlier models, concluding that deep learning algorithms can recognize faces under harsh conditions such as strong occlusion types of illumination, and strong expressions. Nevertheless, recognizing faces in images with low-resolution, extensive pose variations, and open-set recognition still pose challenges. To ensure the reproducibility of their findings, the authors utilized open-source and reproducible software, conducting experiments on six diverse datasets with five different face recognition algorithms. Furthermore, they offered access to their source code to allow others to reproduce the experiments.

In their study, Kai [6] et al. demonstrated the application of deep learning in recognizing human faces using both infrared and visible light images. According to the authors, the model outperformed other state of-the-art approaches, even when faced with variations in illumination.

In their paper, Fatimah [7] et al. proposed a novel approach by integrating the coherence of Discrete Wavelet Transform (DWT) with four distinct algorithms namely: Convolutional Neural Network (CNN), Eigen vector of PCA, error vector of principle component analysis (PCA), and Eigen vector of Linear Discriminant Analysis (LDA). The detection probability entropy and Fuzzy system were utilized to combine the four results. The accuracy of recognition was found to be dependent on the diversity and quality of the image database. The proposed combined method achieved recognition rates of 89.56% and 93.34% for the worst and best cases, respectively, which outperformed previous works where individual methods were implemented on specific image datasets.

The authors of the study, Omkar [8] et al. had the objective of implementing face recognition through the use of a single photograph or a series of faces recorded in a video. They employed a hybrid approach that leveraged both automated and manual methods to construct a large dataset containing 2.6 million images belonging to over 2.6 thousand individuals.

Using SVM techniques, Laith [9] et al. developed facial recognition systems. Image preparation was the first step. They used the Contrast stretching and Normalization size of the picture procedures. The image was then reduced to half its original size, the noise was removed, the processing time was cut in half, and features were extracted for classification.

A novel CNN architecture was presented by S. Meenakshi [10] et al. to eliminate the impact of changes in position and lighting, any occlusions, facial emotions, etc. Convolutional layers C1, C2, and C3 are used in the created approach to experiment with different feature maps in an effort to identify the most effective architecture. Moreover, this architecture contains a fully connected layer, a subsampling layer, and an input layer with a 32 x 32 pixel picture. After shrinking the image to 32 by 32 pixels, numerous tests are run on the ORL database to gauge the model's effectiveness. The 15-90-150 design has the highest accuracy of all the architectures, at 98.75%.

Putta [11] et al. employed PCA for feature reduction and extracted rotation and scale invariant characteristics from the normalized facial picture using the Gabor Wavelet. The classification was carried out by these three writers using SVM. The accuracy rates obtained on the three data sets: ORL, AR, and Grimace, respectively: 97.65%, 92.31%, and 100.00%.

The PCA approach was utilized by Ni [12] et al. to extract distinguishing characteristics. They also used gray scale conversion, region of interest (ROI), and Haar Cascade Segmentation for picture preprocessing. After that, the KNN algorithm is used to classify the data. While PCA+KNN approach accomplished an 81% recognition rate on a dataset research comprising 790 faces from 158 persons collected from various perspectives, 2D-PCA+KNN method achieved an accuracy rate of 96.88% on ORL database.

Zhiming [13] et al. proposed a facial recognition model. This model is based on the quantity of convolutional layer feature maps and the quantity of hidden layer neurons. As a result, facial recognition's accuracy has increased. The input layer, convolution layer 1, pooling layer 1, convolution layer 2, pooling layer 2, fully connected layer, and Softmax regression classification layer are all parts of the CNN architecture. As a result, they established the structure C1-C2-H, where C1 denotes the quantity of feature mappings in the first convolutional layer, C2 the quantity in the second convolutional layer, and H the quantity of hidden layer neurons. The ideal model, 36-76-1024, was discovered by Zhiming [13] et al. through numerous experimental experiments. Also, they achieved a facial recognition rate of 100% on the ORL database.

Throughout 2021, a team of researchers called Muhammad [14] et al, via their experimental work, undertook a comparative analysis of four classical algorithms for machine learning using the ORL database. The comparison includes PCA, 1-Nearest Neighbor (1-NN), LDA, and SVM as the four machine learning methods. Then, the researchers developed the models, trained the classifiers, and retrieved features from the datasets. Finally, they used the 5-fold cross validation (n=5) method to assess how well these models performed. The accuracy ratings for the systems based on LDA, 1-NN, PCA, and SVM were 96%, 96.25%, 96.75%, and 98%, respectively. These outcomes show how well the SVM approach for face recognition works.

One of the CNN designs dubbed Residual Networks-50 was utilized by Yohanssen [15] et al. in their research to produce a system for recognizing faces. For the ImageNet test set, these cnn models achieve an error rate of 3.57%. The contribution of this research study is to establish efficacy ResNet architecture utilizing various configurations of hyper parameters such as the amount of hidden layers, the amount of units included in the hidden layer, number of iterations, and learning rate. Dataset size of 1050 photos split into the training and testing sets, with a ratio of 80% for the train dataset and 20% for the test dataset, to evaluate the model. They discovered that a learning of 0.0001, an epoch size of 100, and a step size of 150 result in a model with a 99% accuracy rate.

A comparison of machine learning techniques in the field of facial recognition was conducted by Benradi [16] et al. These researchers acquired photographs from two databases: the ORL and the Sheffield face databases, which have 564 images of 20 people with identical dimensions of 220 220 pixels and 256-bit grayscale. They applied feature extraction utilizing the Scale-Invariant Feature Transform (SIFT), the Speeded Up Robust Features (SURF), the Features from Accelerated Segment Test (FAST), and LBP to these databases' two datasets, which were partitioned into train and test datasets. SVM, KNN, PCA and 2D-PCA algorithms are used in classification to develop prediction models using the feature vectors collected from the face photos. The forecasting models were examined, the outcomes demonstrated that the suggested approaches, including SIFT+SVM, LDA+KNN, PCA and 2D-PCA performed with the following accuracy rates in the ORL Dataset: 99.16%, 96%, 92.50% and 96.25%. The accuracy ratings for the algorithms SIFT+SVM, LDA+KNN, PCA and 2D-PCA on the Shiefilled Dataset are 99.44%, 96%, 27.11% and 43.10% respectively. Definitely SVM out performs the other investigated algorithms, according to Benradi [16] et al.

The two algorithms will be combined in 2022 in order to develop a facial recognition system, according to Zahraa [17] et al. The SVM method is the first one. The second is a brand-new meta heuristic technique named Rain optimization algorithm (ROA), which was motivated by rainfall. If its parameters are properly adjusted, this approach can find both local and global extremism. The aim of the Radial Basis Function (RBF) kernel SVM's C and parameters optimization in this work is to use ROA. They therefore were using the Yale face dataset to assess the proposed approach. By carrying out an n-fold validation (n = 10) the authors were able to achieve an 86% identification rate.

A paper by Safa [18] et al. explores the application of machine learning algorithms in healthcare, particularly for anemia disease classification. It underscores the significance of processing large healthcare datasets using digital analytics and classification tools. The study conducts a comparative analysis of twelve classification algorithms, revealing that Logitboost, Random Forest, XGBoost, and Multilayer Perceptron performed well, with XGBoost achieving the highest accuracy. Subsequently, XGBoost is employed for classifying new datasets in Hematology studies in Iraq.

Connecting this with machine learning algorithms used for face recognition, the paper highlights the broader utility of machine learning techniques in healthcare and beyond. Just as these algorithms effectively classify medical conditions, they can also be harnessed for tasks like face recognition, offering promising results. However, it's crucial to ensure high-quality and diverse training data for both healthcare and face recognition applications, as well as acknowledging potential errors, especially in complex scenarios. Overall, this paper underscores the significant potential of machine learning algorithms in enhancing medical diagnoses and treatments and suggests their relevance in various domains, including face recognition.

These studies demonstrate the potential of machine learning algorithms to recognize faces. According to our knowledge, there is not dataset specific to Arabic actor's images; and there is not a system that can find out movie's year of production from actors ages in that movie. So our study adds to this body of work by building a system that uses machine learning to classify and recognize faces of Arabic actors from images taken randomly from Arabic movies and calculating the year of movie's production. To increase performance of the system a comparison of different machine learning algorithms was made. The results of our study can be used to develop more accurate and reliable face recognition systems for various applications.

## 3. Methodology:
### 3.1. Arabic Actor Dataset-AAD Collection and Analysis:

Face picture databases are required for the deployment and evaluation of a facial recognition system. Table 1 shows some of the options that have a lot of interest among researchers where whole image collection was captured under various facial expression conditions.

Table 1. -Some Face picture databases.

| Database Name | No. of Pictures | No. of Persons | Image Resolution | Lightening |
|---|---|---|---|---|
| GRIMACE[11] | 360 | 18 | 200×180 pixels | identical lighting conditions |
| ORL[11,19] | 400 | 40 | 112 × 92 pixels | includes subtle differences in illumination, in pose, and face characteristics |
| YALE[20] | 165 | 15 | ---- | different illumination |

In Table 1 images were identical in resolution for each system; and only a face occur in each image as in Fig.1.

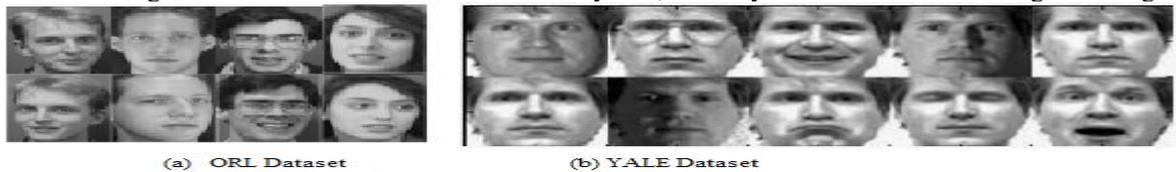

Figure 1. Sample images. (a) ORL dataset. (b) YALE Dataset.

In this research a new dataset was built by collecting 574 images from 14 different Arabic movies for 14 actors (7 women and 7 men) as shown in Table 2 and Table 3. These images were divided into two separate sets, the training and testing sets. The model will be trained using the first one, and evaluated using the second. Training set consisted of 468 images; while testing set had 106 images.

The use of images taken from Arab movies includes dealing with several challenges, such as non-uniform lighting in the images, Different and multiple poses for the actors, One image contains several elements (such as the furniture in the room) with the actor or a group of actors, and not as previous researches for face classification and recognition used to deal with images that include the face of the person only, the use of make-up, wigs, beards, and wearing different accessories and costumes commensurate with the type of role played by the actor (for example, the actress Madiha Kamel in some pictures wearing the Egyptian Abaya, the Egyptian veil, and the Egyptian Niqap) made it difficult for the system to identify the personality of the same actor, and different age stages were chosen for the same actor (for example, Nour al-Sharif. A picture was taken from a movie; he was twenty years old, with other pictures of the same actor from another movie, and he is forty years old). Also, the images taken from various movies, including black and white, and color ones (old and new movies). Also some images represent full scene while others were part of a scene. In addition, as far as we know, there is no dataset dedicated to images taken from Arab films, so we have created and developed one (Arabic Actors Dataset-AAD) utilizing an 82/18 split ratio for the training and testing sets. Fig.2 shows sample images from AAD.

Table 2. -System Database (14 movies, 14 actors, 574 images).

| No. | Movie Title | Production Year |
|---|---|---|
| 1 | من البيت للمدرسة From Home To School | 1972 |
| 2 | عندما يبكي الرجال When Men Cry | 1984 |
| 3 | مع سبق الاصرار Premeditation | 1979 |
| 4 | شلة المحتالين Bunch Of Crooks | 1973 |
| 5 | النصابين Swindlers | 1984 |
| 6 | انا Me | 1985 |
| 7 | نساء في المدينة Women In City | 1977 |
| 8 | شوارع من نار Streets Of Fire | 1984 |
| 9 | دعوني انتقم Let Me Revenge | 1979 |
| 10 | غروب و شروق Sunset And Sunrise | 1970 |
| 11 | الراقصة والطبال Dancer And Drummer | 1984 |
| 12 | بنات ابليس Satan's Daughters | 1984 |
| 13 | لعبة الانتقام Revenge Game | 1992 |
| 14 | صراع الاحفاد The Struggle of the Grandchildren | 1989 |

Table 3. -Actors with their ages.

| No. | Actor Name | Birth Date | Death Date |
|---|---|---|---|
| 1 | نبيلة عبيد Nabila Aubaid | 1945 | --- |
| 2 | محمود ياسين Mahmood Yasin | 1941 | 2015 |
| 3 | نور الشريف Nour al-Sharif | 1946 | 2015 |
| 4 | محمود المليجي Mahmood Al-Mileegy | 1910 | 1983 |
| 5 | ليلى علوي Layla Alui | 1962 | --- |
| 6 | مديحة كامل Madiha Kamel | 1948 | 1997 |
| 7 | سيد زيان Sayed Zayan | 1943 | 2016 |
| 8 | حسين فهمي Husain Fahmi | 1940 | --- |
| 9 | ميرفت امين Mirvat Ameen | 1948 | --- |
| 10 | فريد شوقي Fareed Shawqi | 1920 | 1998 |
| 11 | شويكار Shwikar | 1936 | 2020 |
| 12 | سعاد حسني Suad Husni | 1943 | 2001 |
| 13 | اسعاد يونس Isaad Younis | 1950 | --- |
| 14 | توفيق الذقن Twfiq Al-Thiqin | 1923 | 1988 |

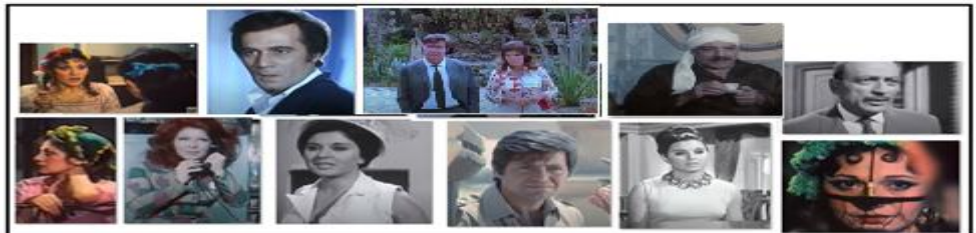

Figure 2. Sample images from AAD.

## 3.2. The Method:

Complete system components and workflow is shown in Fig. 3 below. ADD was prepared and divided for training and testing. Image feature extraction was done for both training and testing sets. Then machine learning algorithms were used for the classifier and predictor. Performance evaluation was made to select the most accurate algorithm to be used to predict the movie production year for new images with known attributes. Fig. 4 is an algorithmic representation of the methodology.

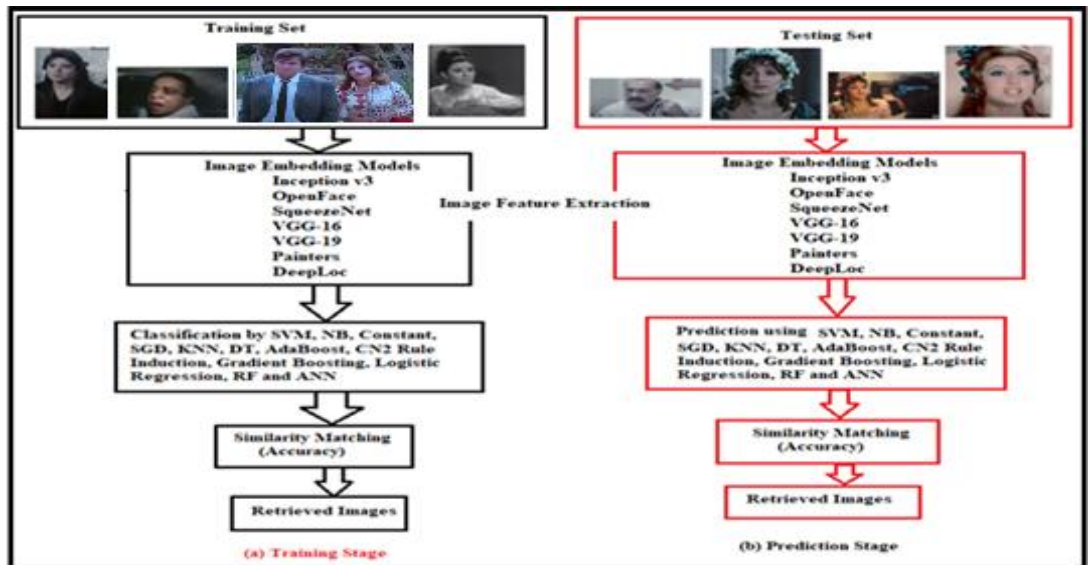
Figure 3. System Components and Workflows.

1. **Arabic Actor Dataset (AAD) Creation:**
   - Collect 574 images from 14 different Arabic movies featuring 14 actors.
   - Divide the dataset into training (468 images) and testing sets (106 images).

2. **Image Feature Extraction:**
   - Initialize a list to store the extracted features.
   - For each image in the AAD:
     - Apply image embedding models (Inception v3, OpenFace, SqueezeNet, VGG-16, VGG-19, Painters, and DeepLoc) to obtain low-dimensional representations.
     - Add the extracted features to the list.

3. **Classifier Training:**
   - Initialize the training set features and labels.
   - For each image in the training set:
     - Assign the corresponding image features from the extracted features list.
     - Assign the label based on the actor's identity.
   - Train machine learning algorithms (SVM, NB, Constant, SGD, KNN, DT, AdaBoost, CN2 Rule Induction, Gradient Boosting, Logistic Regression, RF, and ANN) using the training set features and labels.

4. **Face Recognition:**
   - Initialize the testing set features and predicted labels.
   - For each image in the testing set:
     - Assign the corresponding image features from the extracted features list.
     - Use the trained machine learning algorithms to predict the actor's identity.
     - Store the predicted label.

5. **Movie Production Year Calculation:**
   - For each image in the testing set:
     - If the predicted label matches the actual label:
       - Use the predicted movie production year from the algorithm as the output.
     - If the predicted label does not match the actual label:
       - Compute the movie production year using the known actor's birth year and their age at the time of the image.

6. **Performance Evaluation:**
   - Calculate evaluation metrics (AUC, accuracy, F1 score, precision, recall, log loss, and specificity) for the trained classifier and prediction models.
   - Compare the performance of different machine learning algorithms.
   - Identify the most accurate algorithms for actor identification and movie production year prediction.

Figure 4. Algorithmic Representation of the Methodology.

## 3.3. Feature Extraction:

From the foregoing, images with similar specifications that were mentioned; represent a great challenge for image embedding models which is performed to obtain a low-dimensional representation of a given image that accurately captures its key elements and allows for straightforward comparison with other embedding's [21.22].

There are several face recognition models available for free that can be used for various applications such as OpenCV [23,24], Face Recognition, OpenFace [25], FaceNet [26], DeepFace [27] and DLib [24] Face Recognition. The best model for your application will depend on your specific requirements and constraints.

In this research several image embedding models were used such as Inception v3, OpenFace, SqueezeNet, VGG-16, VGG-19, Painters and DeepLoc. A general summary on each model is shown in Table 4 below. The N/A means that there is no top-1 or top-5 accuracy information available for the model.

Table 4. -A Comparison of Image Embedding Models.

| Model | Description | Number of parameter | Top-1 accuracy % | Top-5 accuracy % | Image Type |
|---|---|---|---|---|---|
| Inception v3[28] | A deep neural network designed for image recognition, with improved efficiency over previous Inception models. | 23.8 million | 78.8% | 94.4 | General Image Classification |
| OpenFace[29] | A face recognition model that uses deep neural networks to extract features from faces, designed to work under different conditions | 2.3 million | 92.2% | 94. | Face Recognition |
| SqueezeNet[30] | A lightweight neural network designed for image classification with low memory and computational requirements | 0.72 million | 57.5% | 80.3 | General Image Classification |
| VGG-16[31, 32, 33] | A deep neural network with 16 layers, designed for image recognition with a focus on improving accuracy | 138.3 million | 71.5 | 90.2 | General Image Classification |
| VGG-19[34] | A deeper version of VGG-16 with 19 layers, designed to improve accuracy further | 143.7 million | 72.3 | 90.8 | General Image Classification |
| Painters[35] | A neural network designed for image classification that can distinguish between paintings by different artists | 16.6 million | 80.2 | N/A | Painting Classification |
| DeepLoc[36] | A neural network designed for protein localization prediction in microscopy images | 1.1 million | 85.7 | N/A | Protein Subcellular Localization |

## 3.4. Classifier Training:

Machine learning algorithms like (SVM, NB, Constant, SGD, KNN, DT, AdaBoost, CN2 Rule Induction, Gradient Boosting, Logistic Regression, RF and ANN) were used to build a model that can classify and recognize actor's faces. Table 5 shows a general comparison among these algorithms. Where in Constant algorithm predict the most frequent class or mean value from training set; this is related to baseline models in machine learning[32].

Table 5. -Machine Learning Algorithms Compared for Face Classification and Recognition.

| Algorithms | Advantages | Disadvantages | Suitable Application |
|---|---|---|---|
| SVM[37,51] | Works well on high-dimensional data | Can be sensitive to the choice of kernel function | Face recognition, classification |
| NB[38, 39, 40] | Simple and easy to implement | Assumes independence of features, which may not hold in practice | Face classification |
| Constant[41] | Simple and interpretable | May not perform well on complex data | Face classification |
| SGD[42] | Efficient for large-scale datasets | May not converge to a global minimum | Face classification |
| KNN[12] | Simple to implement and interpret. Good for small datasets, works well with local patterns. | Can be computationally expensive for large datasets. May not work well with high-dimensional data. | Face classification |
| DT[43] | Easy to understand and interpret. Can be effective in classification tasks with clear decision boundaries. | Prone to over fitting, especially with complex datasets. May not work well with continuous data. | Classification or Recognition |
| AdaBoost[44] | Can achieve high accuracy with multiple weak classifiers. Good for face recognition tasks. | Prone to over fitting if weak classifiers are too complex. Can be computationally expensive. | Recognition |
| CN2 Rule Induction[45] | Can handle complex datasets and produce interpretable rules. Can be effective in face classification tasks. | May not work well with small datasets. Can be sensitive to noise and outliers. | Classification |
| Gradient Boosting[46] | Can achieve high accuracy with multiple weak learners. Good for face recognition tasks. | Can be computationally expensive for large datasets. Prone to over fitting if weak learners are too complex. | Recognition |
| Logistic Regression[47, 48] | Simple to implement and interpret. Can be effective in face classification tasks when using binary outcomes. | May not work well with nonlinear relationships between features and outcomes. | Classification |
| RF[49] | Can achieve high accuracy with multiple decision trees. Good for face recognition tasks. | Can be computationally expensive for large datasets. May not work well with imbalanced data. | Recognition |
| ANN[20] | Highly effective in face recognition tasks when trained with large datasets. | Can be expensive to compute and needs a lot of training data. And is prone to fitting too tightly. | Recognition |

### 3.5. Face Recognition:

This is done by using test set. In this stage; the output of image embedding is the input to prediction model. Prediction is done using learning algorithms like (SVM, NB, Constant, SGD, KNN, DT, AdaBoost, CN2 Rule Induction, Gradient Boosting, Logistic Regression, RF and ANN) to select the most accurate result among them. Confusion matrix was used to summarize the predictions made by a model on a set of test data, comparing them to the actual values or labels of the data. The matrix is constructed as follows: (1) the rows of the matrix correspond to the actual or true labels of the data. (2) The columns of the matrix correspond to the predicted labels of the data. (3) Each cell in the matrix represents the number of times a data point was classified or predicted as belonging to a certain class, given its true label.

### 3.6. Calculate Movie Production Year:

If the image is classified correctly by the machine learning algorithm, use the predicted movie production year from the algorithm as the output. Else if the image is misclassified, compute the movie production year using Eq. 1:

$$\text{Movie Production Year} = \text{Actor Birth Year} + \text{Age of Actor} \quad \ldots\ldots\ldots.1$$

Where the Actor Birth Year is assumed to be known.

## 4. Results and Discussion:

Table 6 contains a comparison of the results obtained from using image embedding models for image feature extraction. Despite the fact that OpenFace was specifically designed for face recognition but it had the worst result in image embedding for ADD. VGG-19 had a run time error. While other models (Inception v3, SqueezeNet, Painters, and DeepLoc and VGG-16) had identical results thus Inception v3 was used to train the classifier and predictor.

Tables 7 shows a comparison of results obtained from using these machine learning algorithms in training the classifier when using Inception v3 embedding model. The table shows Training Time, Test Time, AUC-ROC, Accuracy, F1, Precision, Recall, LogLoss, and Specificity.

Table 6. A Comparision of Results from Image Embedding Models.

| Model Name | Training (Images successfully embedded) | Training (Skipped Images) | Testing (Images successfully embedded) | Testing (Skipped Images) |
|---|---|---|---|---|
| Inception v3 | 468 | 0 | 106 | 0 |
| SqueezeNet | 468 | 0 | 106 | 0 |
| VGG-16 | 468 | 0 | 106 | 0 |
| VGG-19 | Run time error | 468 | Run time error | 106 |
| Painters | 468 | 0 | 106 | |
| DeepLoc | 468 | 0 | 106 | 0 |
| OpenFace | 349 | 119 | 89 | 17 |

Table 7. Results from Training Classifier.

| Model | Train time [s] | Test time [s] | AUC | CA | F1 | Precision | Recall | LogLoss | Specificity |
|---|---|---|---|---|---|---|---|---|---|
| Logistic Regression | 69.901 | 5.652 | 0.990 | 0.855 | 0.842 | 0.860 | 0.855 | 0.431 | 0.973 |
| Neural Network | 50.885 | 10.841 | 0.988 | 0.846 | 0.835 | 0.830 | 0.846 | 0.551 | 0.976 |
| SGD | 27.283 | 11.391 | 0.906 | 0.838 | 0.821 | 0.812 | 0.838 | 5.609 | 0.983 |
| SVM | 41.302 | 13.183 | 0.985 | 0.797 | 0.765 | 0.786 | 0.797 | 0.698 | 0.944 |
| kNN | 9.001 | 9.986 | 0.958 | 0.767 | 0.733 | 0.727 | 0.767 | 1.824 | 0.960 |
| Gradient Boosting | 2364.893 | 6.130 | 0.913 | 0.684 | 0.675 | 0.671 | 0.684 | 2.227 | 0.949 |
| Random Forest | 10.972 | 4.851 | 0.903 | 0.652 | 0.619 | 0.609 | 0.652 | 2.882 | 0.929 |
| Tree | 100.377 | 0.128 | 0.742 | 0.534 | 0.529 | 0.539 | 0.534 | 14.725 | 0.931 |
| AdaBoost | 13.689 | 6.186 | 0.718 | 0.515 | 0.516 | 0.518 | 0.515 | 16.753 | 0.933 |
| CN2 rule inducer | 33367.200 | 5.685 | 0.755 | 0.476 | 0.475 | 0.477 | 0.476 | 1.997 | 0.927 |
| Constant | 0.015 | 0.051 | 0.491 | 0.267 | 0.113 | 0.071 | 0.267 | 2.256 | 0.733 |
| Naive Bayes | 21.653 | 8.757 | | 0.060 | 0.065 | 0.238 | 0.060 | | 0.989 |

## 5. Performance Evaluation:

Evaluate the Performance includes; (a) Evaluate the classifier; after training system classifier, evaluations of different embedding models and machine learning algorithms were performed to see their performance on the training set using metrics like Area Under Curve (AUC), Precision, Classification Accuracy (CA), F1 score, specificity and LogLoss. If accuracy used as a measure for performance then the best algorithm used for classification is Logistic Regression 85.5 % followed by Neural Network algorithm with accuracy 84.6%. (b) Evaluate predictor; the confusion matrix allows us to calculate several metrics that are commonly used to evaluate the performance of a classification model, including accuracy, precision, recall, and F1 score. These metrics help us to understand how well the model is able to correctly identify different classes of data, and to diagnose any specific areas where the model may be performing poorly[50]. Fig. 5 shows confusion matrix for Logistic Regression; where blue cells used for correct result with its ratio and pink color used for wrong result with its ratio. This figure shows prediction results when using test set including 106 random selected images.

Figure 5. Confusion Matrix for Logistic Regression Learning Algorithm.

A comparison of the results obtained from using different machine learning algorithms shown in Table 8 is performed. Both Logistic Regression and SGD has the highest equal correct results thus they can be considered as the best machine learning algorithms used for prediction for images taken randomly from different movies.

Table 8. -Results of Different Machine Learning Algorithms in Prediction.

| No. | Algorithm Name | Correct Prediction | Wrong Prediction |
|---|---|---|---|
| 1 | Logistic Regression | 85 | 21 |
| 2 | Neural Network | 80 | 26 |
| 3 | SGD | 85 | 21 |
| 4 | SVM | 82 | 24 |
| 5 | Random Forest | 71 | 35 |
| 6 | AdaBoost | 53 | 53 |
| 7 | Gradient Boosting | 80 | 26 |
| 8 | KNN | 80 | 26 |
| 9 | CN2 rule inducer | 50 | 56 |
| 10 | Tree | 47 | 59 |
| 11 | Constant | 25 | 81 |
| 12 | Naïve Bayes | 11 | 95 |

According to results obtained from Table 8; selection of best performing algorithm to be used to predict the movie production year for new images with known attributes.

## 6. Conclusions:

Using images from Arabic movies, this paper concluded with a comparative review of several image embedding models and machine learning techniques for face identification. Different lighting situations, actor positions, and the presence of several items in the scenes were just a few of the difficulties that the images brought. Furthermore, the use of makeup, accessories, and costumes made it harder to distinguish between performers, especially when they were portraying different age groups. Despite these challenges, our analysis revealed valuable insights. Comparing a number of image embedding models, it was discovered that Inception v3 produced the best trustworthy findings for feature extraction. We then assessed several machine learning age prediction and actor categorization strategies. During the training phase, Logistic Regression performed better than other methods, displaying high AUC, precision, classification accuracy, and F1 score values. Both Logistic Regression and SGD showed to be reliable prediction algorithms throughout the testing phase.

This research has significant implications for applications such as movie search engines, recommendation systems, and genre analysis. This research helps to comprehend changing film industry trends and offers insightful information on human behavior and demography, particularly as it relates to aging. This is done through enhancing the accuracy and dependability of facial recognition systems.

Despite the limitations of our study, which include a restricted dataset consisting only of Arabic movies and the inherent challenges in identifying actors, our research serves as a stepping stone for future advancements in face recognition systems. Future research should broaden their datasets to include bigger and more varied populations in order to improve the generalizability of their findings. Additionally, including other elements like facial expressions and utilizing deep learning methods might help to increase the accuracy of age prediction. These research directions show potential for improving facial recognition technologies and broadening their uses outside the purview of our current study.

# تحليل مقارن لخوارزميات التتعلم الالي للتعرف على الوجوه للعثور على سنة انتاج الفلم

**الخلاصة:**

استخدمت هذه الدراسة خوارزميات تعلم الآلة المختلفة لتحديد والتعرف على الممثلين واستخراج عمر الممثلين من الصور المأخوذة عشوائيًا من الأفلام العربية. ومن ثم احتساب سنة انتاج الفلم من اعمار ممثليه التي توصل اليها النظام. يتضمن استخدام الصور المأخوذة من الأفلام العربية تحديات مثل الإضاءة غير الموحدة ، وطرح مختلف ومتعدد للممثلين وعناصر متعددة مع الممثل أو مجموعة من الممثلين. بالإضافة إلى ذلك ، فإن استخدام الماكياج والشعر المستعار واللحية وارتداء الملحقات والأزياء المختلفة الخاصة بالشخصية التي يؤديها الممثل جعل من الصعب على النظام تحديد شخصية الممثل نفسه (الشخصية الحقيقية). تم اختيار مراحل عمرية مختلفة للممثل نفسه ، مثل نور الشريف التي تبلغ من العمر عشرين عامًا في احد الافلام وأربعين عامًا في فلم اخر. وتم انشاء مجموعة بيانات الممثلين العرب هي مجموعة بيانات من 574 صورة مأخوذة من أفلام مختلفة ، بما في ذلك افلام الأسود والأبيض والافلام الملونة. كما تمثل بعض الصور مشهدًا كاملاً بينما كان البعض الآخر جزءًا من مشهد. تمت المقارنة بين عدة نماذج لاستخراج الميزات. كما تم إجراء مقارنة بين مجموعة من خوارزميات التعلم الآلي المختلفة في مرحلتي التصنيف والتنبؤ لمعرفة الخوارزمية الأفضل في التعامل مع مثل هذا النوع من الصور. أظهرت الدراسة فعالية نموذج الانحدار اللوجستي الذي أظهر أفضل أداء مقارنة بالنماذج الأخرى في مرحلة التدريب ، كما يتضح من قيم المنطقة تحت المنحنى (AUC) والضبط و الدقة و F1 التي بلغت 99٪ و 86٪ و 85.5٪ و 84.2٪ على التوالي. يمكن استخدام نتائج هذا البحث لتعزيز الدقة والاعتمادية لأنظمة التعرف على الوجه لتطبيقات متنوعة مثل محركات البحث عن الأفلام وأنظمة التوصية بالأفلام وتحليل نوع الأفلام.

**الكلمات المفتاحية:** الذكاء الاصطناعي ، وخوارزميات التعلم الآلي ، والتعرف على الوجوه ، والتنبؤ بالعمر، نايف بايسين، شجرة القرار ، آلة المتجهات الداعمة ، والشبكة العصبية الاصطناعية.